\documentclass{article}
\usepackage{spconf,graphicx}
\usepackage{amsmath,amsfonts}
\usepackage{algorithmic}
\usepackage{algorithm}
\usepackage{array}
\usepackage{textcomp}
\usepackage{stfloats}
\usepackage{url}
\usepackage{adjustbox}
\usepackage{verbatim}
\usepackage{graphicx}
\usepackage{epstopdf}
\usepackage{cite}
\usepackage{booktabs}
\usepackage{hyperref}
\hypersetup{hidelinks}

\title{SPATIO-TEMPORAL CO-ATTENTION FUSION NETWORK FOR VIDEO SPLICING LOCALIZATION}
\name{Man Lin$^{1,2}$, Gang Cao$^{1,2}$$^{\ast}$ \thanks{*Corresponding author}, Zijie Lou$^{1,2}$}
\address {\normalsize{$^{1}$State Key Laboratory of Media Convergence and Communication, Communication University of China, Beijing 100024, China}\\
\normalsize{$^{2}$School of Computer and Cyber Sciences, Communication University of China, Beijing 100024, China}}

\begin{document}
%
\maketitle
\begin{abstract}
Digital video splicing has become easy and ubiquitous. Malicious users copy some regions of a video and paste them to another video for creating realistic forgeries. It is significant to blindly detect such forgery regions in videos. In this paper, a spatio-temporal co-attention fusion network (SCFNet) is proposed for video splicing localization. Specifically, a three-stream network is used as an encoder to capture manipulation traces across multiple frames. The deep interaction and fusion of spatio-temporal forensic features are achieved by the novel parallel and cross co-attention fusion modules. A lightweight multilayer perceptron (MLP) decoder is adopt-ed to yield a pixel-level tampering localization map. A new large-scale video splicing dataset is created for training the SCFNet. Extensive tests on benchmark datasets show that the localization and generalization performances of our SCFNet outperform the state-of-the-art. Code and datasets will be available at \href{https://github.com/multimediaFor/SCFNet}{https://github.com/multimediaFor/SCFNet}.  
\end{abstract}
\begin{keywords}
 Video forensics, digital video forgery, video splicing localization, co-attention fusion
\end{keywords}
\section{Introduction}
With rapid development of digital media editing techniques, digital video manipulation becomes rather convenient and easy. Malicious users might forge a video which is difficultly distinguished from authentic ones. The object-based forgeries, such as object insertion and removal, pose great threats\cite{richao2014detection}. The object insertion splices spatio-temporal regions from the same or different videos. It significantly alters the semantic content, such as persons, vehicles and logos. Misuse of such video forgeries incur serious implications on legal matters, public opinion and intellectual property rights. In comparison to images, the blind detection of video splicing is challenging due to the extra motion disturbance and higher data dimensions\cite{jin2021object}. As pointed out by the prior work\cite{nguyen2022videofact}, video compression inevitably weakens the forensic traces, resulting in the failure of image forensic techniques. As a result, it is significant to develop specialized forensic techniques for video splicing localization.

In recent years, there are substantial achievements on image tampering localization techniques\cite{cozzolino2019noiseprint, dong2022mvss, liu2022pscc}. In contrast, video forensics have received less attention. The traditional artificial feature approaches rely on detecting the artifacts left by compression\cite{labartino2013localization}, imaging pipeline\cite{kobayashi2010detecting} and interframe motion\cite{kancherla2012novel}, etc. To discover anomalous videos, Avino \textit{et al}.\cite{d2017autoencoder} and Johnston \textit{et al}.\cite{johnston2020video} leverage autoencoders to learn the intrinsic model of genuine videos. However, the applicability of such methods is limited and merely suited to specific video contents and formats. Due to the absence of large-scale video forgery datasets, \cite{jin2021object, jin2022video} use image datasets to train the video forensic networks. Video object tracking and segmentation are used to temporally refine the preliminary splicing localization maps within single frames. Although such models improve the localization accuracy, their robustness against post compression is still limited. In the latest VideoFact scheme\cite{nguyen2022videofact}, the network with contextual embedding is designed to enhance the robustness against post video compression. Although it addresses various types of video forgeries, both the constructed training dataset and the forensic network merely exploited single-frame information. The temporal correlation between frames is ignored, resulting in unfavorable generalization ability.

To attenuate the deficiencies of prior works, the spatio-temporal co-attention fusion network (SCFNet) is proposed for effective video splicing localization in an encoder-decoder framework. A three-stream encoder network based on co-attention fusion is proposed to capture the intra-frame and inter-frame tampering traces. The co-attention modules facilitate information exchange among frames for learning discriminative features. Meanwhile, the lightweight All-MLP decoder\cite{xie2021segformer} is adopted to yield binary localization maps.  Besides, we construct a new video splicing dataset (VSD) for training SCFNet. Varying degrees of compression is used as data augmentation to improve the robustness of our model. Effectiveness of SCFNet is verified by extensive assessments on public benchmark datasets. 

\section{PROPOSED SCHEME}

Overall network architecture of our SCFNet is shown in Fig. 1. For the three-stream encoder network, modified Res-Net101\cite{he2016deep} is used as backbone, which is fed with three consecutive frames of input videos. Co-attention modules are integrated to the encoder for fusing the features of different frames. The lightweight All-MLP decoder is used to aggregate multiscale features and yield pixel-level localization maps. Details of the scheme are elaborated below.

\subsection{Three-stream encoder via co-attention fusion}
The video splicing process involves replacing certain regions in a video with the content from other videos. Regions from different videos inherently exhibit inconsistent distribution of texture, noise and compression artifacts. Such inconsistencies are often observable in the temporal domain as well. Thus, three consecutive frames $\mathrm{I}_{t-1}$, $\mathrm{I}_{t}$, $\mathrm{I}_{t+1}$ are input to the three-stream encoder, enabling the network to explore spatio-temporal anomalies. Each branch is built on ResNet101,  which is composed of a convolutional layer and four stages stacked with multiple residual blocks. In the stages 2, 3, 4, features are first extracted separately for each branch, then merged and enhanced by co-attention fusion modules. The resulting intermediate features, which have undergone spatio-temporal information interaction, are forwarded to the next stage for extracting deeper features. Furthermore, they are integrated through the global-local attention context module (GAC) \cite{pei2023hierarchical} to serve as input for the decoder. The workflow of the encoder is shown in the upper part of Fig1.

Inspired by\cite{pei2023hierarchical, jaderberg2015spatial}, co-attention mechanism is adopted to fuse the spatio-temporal features of neighboring frames. It captures the interaction among multiple features, and suitable for the case of symmetrical features. The involved parallel and cross co-attention fusion modules are described below.

\begin{figure*}[htbp]
\centering
\includegraphics[width=\textwidth]{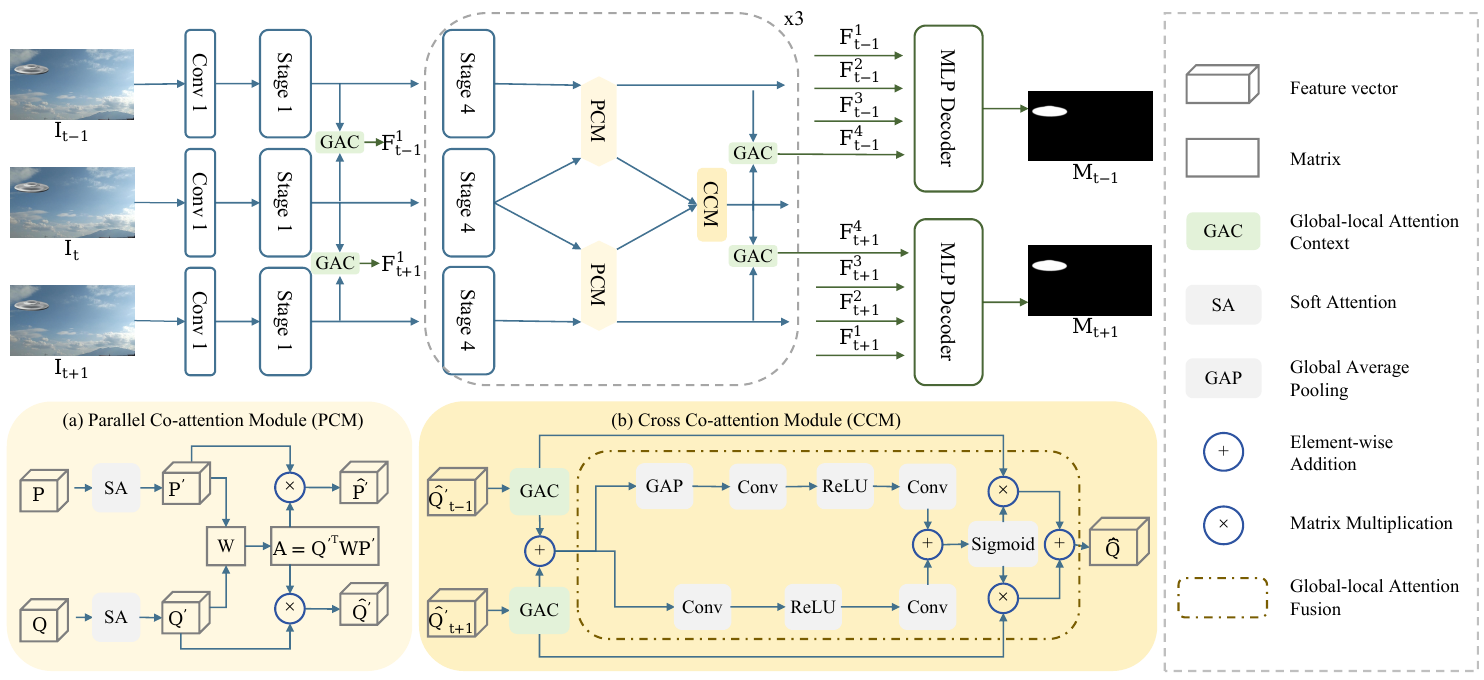}
\caption{Overall pipeline of the proposed video splicing localization network SCFNet.}
\label{fig_1}
\end{figure*}

\noindent 
\textbf{Parallel co-attention module. }As illustrated in Fig. 1 (a), intermediate features from adjacent branches  $\mathrm{P} \in \mathbb{R}^{W \times H \times C}$ and $\mathrm{Q} \in \mathbb{R}^{W \times H \times C}$ are converted to enhanced preliminary attention features $\mathrm{P}^{'}$ and $\mathrm{Q}^{'}$ by soft attention. Then the co-attention method\cite{lu2019see} is used to compute a correlation matrix $\mathrm{A} \in \mathbb{R}^{WH \times WH}$ for achieving mutual enhancement and fusion of two feature maps. This co-attention process can be formalized as follows:

\begin{align}
&\mathrm{A}=\eta\left({\mathrm{Q}^{'}}^{\mathrm{T}}\right)\mathrm{W}\eta\left(\mathrm{P}^{'}\right) \nonumber \\ 
&\hat{\mathrm{P}^{'}}= \mathrm{P}^{'}\mathrm{S}_{row}\left(\mathrm{A}\right) \tag{1} \\ 
&\hat{\mathrm{Q}^{'}}= \mathrm{Q}^{'}\mathrm{S}_{col}\left(\mathrm{A}\right) \nonumber
\end{align}

\noindent 
where $\eta$ denotes a linear transformation that maps the features to a lower-dimensional space and $\mathrm{W} \in \mathbb{R}^{C \times C}$ is a learnable weight matrix. $\mathrm{S}_{row}$  and $\mathrm{S}_{col}$ indicate normalization through row-wise and column-wise softmax operations, respectively. 

\noindent 
\textbf{Cross co-attention module. }After fusing information between two adjacent frames by the PCM, the CCM is further utilized to facilitate information exchange among three consecutive frames. It consists of the GAC module and the global-local attention fusion (GAF) \cite{pei2023hierarchical} module. They perform global-local attention enhancement on input features and fused features respectively, as shown in Fig. 1 (b). As a result, attention is directed to the following information: local details related to tampering traces and global contextual information associated with video semantics.

\subsection{MLP-based decoder network}
The intermediate branch of the three-stream encoder serves as a bridge for multi-frame information interaction. Consequently, encoders are connected to the first and third branches to obtain their respective localization maps. The lightweight All-MLP decoder \cite{xie2021segformer} consists solely of MLP layers, achieving channel-wise uniformity and fusion of multiple input features. For the multi-scale hierarchical features $\mathrm{F}^i|_{i=1,2,3,4}$ extracted by the encoder, the prediction process for the localization mask M can be formulated as follows:

\begin{align}
 &\hat{\mathrm{F}}^i =\operatorname{Linear}\left(C^i, C\right)\left(\mathrm{F}^\mathit{i} \right), \forall \mathit{i}  \nonumber \\
 &\hat{\mathrm{F}}^i =\operatorname{Upsample}\left(\frac{H}{4} \times \frac{W}{4}\right)\left(\hat{\mathrm{F}}^i \right), \forall \mathit{i}  \nonumber \\
 &\mathrm{F}=\operatorname{Linear}(4C, C)\left(\operatorname{Concat}\left(\hat{\mathrm{F}}^i \right)\right), \forall \mathit{i}  \tag{2} \\
 &\mathrm{M}=\operatorname{Linear}(C, 1)\left(\mathrm{F}\right)  \nonumber \\
 &\mathrm{M}=\operatorname{Upsample}\left(H \times W\right)\left(\mathrm{M} \right) \nonumber
\end{align}

\begin{sloppypar}
\noindent 
where $C_i$ represents the channel number of $\mathrm{F}^i$ and  $\text{Linear}\left(C_{\mathit{in}}, C_{\mathit{out}}\right)(\cdot)$ denotes a linear layer with input and output vector dimensions of $C_{in}$ and $C_{out}$, respectively. 
\end{sloppypar}

\subsection{Video splicing dataset}
Adequate forged video segments are crucial for training deep learning networks. Due to the complexity of video data, existing video tampering datasets are entirely generated manually using multimedia software such as Adobe After Effects. While these synthetic datasets can closely resemble real scenarios, they suffer from limited quantity. As a result, research on video forensics has been confined to training networks on small-scale datasets or image datasets. This can lead to overfitting or the inability to capture the temporal anomalies in forged videos.

In this research, a new video splicing dataset (VSD) is constructed. A total of 2500 videos are initially collected from YouTube-VOS\cite{xu2018youtube} and GOT-10k\cite{huang2019got}. For each video, the pixel-level mask of a specific target object in every frame is obtained semi-automatically. Specifically, we manually annotate the first frame of specific object in each video using EISeg\cite{hao2022eiseg}. Then, masks for all frames are automatically generated based on the initial frame's mask, utilizing the semi-supervised video object segmentation algorithms STCN-VOS\cite{cheng2021rethinking} and Xmem\cite{cheng2022xmem}. A minor amount of manual correction is applied to the generated masks using Xmem to ensure precision. For background videos, we select them from Video-ACID\cite{hosler2019video} which is employed for camera identification, ensuring data source diversity. Lastly, we develop a program to automate the process of splicing foreground objects onto background videos based on the masks, thereby generating spliced videos in bulk.

\begin{table}[!t]
\caption{Details of  test datasets}
\vspace{3pt}
\label{tab1}
\begin{adjustbox}{width=\linewidth}
\begin{tabular}{lcccc}
\toprule                                   
Dataset	&Count	&Format	&Resolution	&Camera Type       \\
\midrule
GRIP\cite{d2017autoencoder} &10	&AVI / MP4	&$1280\times720$	&Dynamic\\
HTVD\cite{singla2023hevc} &10	&MP4	&$1920\times1080$	&Static\\
VideoSham\cite{mittal2023video} &70	&MP4	&$1920\times1080$	&Static \& Dynamic\\
\bottomrule
\end{tabular}
\end{adjustbox}
\end{table}

\begin{table}[!t]
\centering
\caption{Ablation analysis of our proposed SCFNet on the GRIP dataset.}
\vspace{3pt}
\tabcolsep=20 pt
\label{tab2}
\begin{adjustbox}{width=2.5 in}
\begin{tabular}{lcc}
\toprule                                     
Methods       & IoU               & F1                    \\
\midrule
w/o. PCM    & 0.403       & 0.502           \\
w/o. CCM         & 0.418          & 0.518        \\
w/o. GAC          & 0.419          & 0.511       \\
SCFNet          & 0.441          & 0.545        \\
\bottomrule
\end{tabular}
\end{adjustbox}
\end{table}

\section{EXPERIMENTS}

\subsection{Experimental setup}

\noindent 
\textbf{Datasets. }We randomly select a portion of the videos from the constructed VSD for network training and parameter optimization. In total, the training dataset comprises 600 videos with 73.7k frames, while the validation set consists of 50 videos with 5.8k frames. The test dataset consists of spliced videos from three publicly available video tampering datasets: GRIP\cite{d2017autoencoder} , VideoSham\cite{mittal2023video} and HTVD\cite{singla2023hevc}. Details for each is presented in Table 1.

\noindent 
\textbf{Implementation. }The proposed method is implemented using the PyTorch deep learning framework and all experiments are conducted on an A800 GPU. We set each of 3 consecutive frames as an input unit and the batch size is set to 10. All the frames used in training are resized to $512\times512$ pixels. The binary cross-entropy is employed as the loss function and SGD is utilized as the optimizer. The initial learning rate, momentum and weight decay are set as 1e-4, 0.9, 1e-5, respectively. The network is trained for 12 epochs and the learning rate decreases by 50\% every two epochs. Considering the common video post-processing operation,  one quarter of the training dataset is compressed by H.264 with the Constant Rate Factor (CRF) randomly selected from 15, 23 or 30. Such data augmentation greatly improves the robustness of our SCFNet.

\noindent 
\textbf{Evaluation Metrics. }IoU and F1 are utilized as pixel-level metrics. For calculating metrics, thresholding is necessary as the direct outputs of the network are probability values.  Unless otherwise specified, the fixed threshold 0.5 is adopted.

\subsection{Ablation study}
We conducted extensive ablation experiments on the GRIP dataset to analyze how each component of our model contributes to the final localization results. Specifically, three variants are considered: replacing the PCM, CCM and GAC with commonly used element-wise addition operations. The obtained results are shown in Table 2, it can be observed that the improvements in F1 brought by the three modules are 4.3\%, 2.7\% and 3.4\%, respectively. This confirms that these modules effectively fuse and enhance spatio-temporal features in our approach.

\begin{table}[!t]
\caption{Performance comparison with other tampering localization algorithms on different datasets. First ranking is shown in bold. The results obtained at the best threshold are underlined. “-” indicates not available.}
\vspace{3pt}
\label{tab3}
\begin{adjustbox}{width=\linewidth}
\begin{tabular}{lccc}
\toprule                                        
	          &GRIP	&VideoSham	&HTVD           \\
Methods      & IoU/F1               & IoU/F1               & IoU/F1 \\
\midrule
PSCC-Net\cite{liu2022pscc} &0.135/0.243	&0.150/0.228	&0.073/0.134  \\
VideoFact\cite{nguyen2022videofact} &0.112/0.221	&0.001/0.088	&0.000/0.045 \\
SCFNet(ours)	&\textbf{0.441}/\textbf{0.545}	&\textbf{0.241}/\textbf{0.334}	&\textbf{0.124}/\textbf{0.178}   \\
SCFNet(ours)    &\textbf{\underline{0.572}}/\textbf{\underline{0.664}}	& \underline{0.254}/\underline{0.351}	& \underline{0.176}/\underline{0.242} \\
OVFD\cite{jin2021object}        &  \hphantom{-----} -  /\underline{0.641}	&-	            &-          \\
MDFFRL\cite{jin2022video}    & \underline{0.512}/\underline{0.623}  	&-	            &-         \\
\bottomrule
\end{tabular}
\end{adjustbox}
\end{table}

\subsection{State-of-the-art comparison}
For a fair and reproducible comparison, we choose the state-of-the-art that meets one of the following three criteria: 1) pre-trained models released by paper authors, 2) source code publicly available, or 3) experiment results available in papers. In addition, to avoid bias, we only include methods trained on datasets that do not intersect with the test datasets. Accordingly, we choose several published methods as follows:
\vspace{-5pt}
\begin{itemize}
\item{Models available: PSCC-Net\cite{liu2022pscc}. We use these pre-trained models directly.}
\vspace{-5pt}
\item{Code available: VideoFact\cite{nguyen2022videofact}, which we train using author-provided code.}
\vspace{-5pt}
\item{Results available: OVFD\cite{jin2021object} and MDFFRL\cite{jin2022video}. Such methods are evaluated on the GRIP dataset, and we cite their results directly.
}

\end{itemize}

\noindent 
\textbf{Quantitative comparisons. } Quantitative comparison results are summarized in Table 3. OVFD\cite{jin2021object} and MDFFRL\cite{jin2022video} provide localization results under the best threshold. For fairness in comparison, we present the results of our method under the same conditions, as shown in Table 3 with an underline. The results highlight the superior performance our SCFNet across multiple datasets. Specifically, on the highly realistic GRIP dataset, our method achieves remarkable localization performance with 0.441 in IoU and 0.545 in F1. Even on the challenging HTVD dataset, our method continues to perform admirably. Such experimental results demonstrate the remarkable generalization capability of our SCFNet.

\vspace{2pt}
\noindent 
\textbf{Qualitative comparisons. }In addition to the quantitative comparisons, we also compared different methods qualitatively. Fig. 2 shows some examples of localization results. In comparison to other methods, our SCFNet stands out in the precise localization of spliced regions within intricate scenes, despite the occurrence of a few missed regions. 

\noindent 
\textbf{Robustness against compression quality. }To evaluate the robustness of our method against video compression, we conducted comparisons on the GRIP dataset [10]. CRF is a commonly used parameter for controlling compression quality, and 23 is the default value in H.264 encoder. As show in Fig. 3, the localization performance of our SCFNet shows a small drop under H.264 compression. Specifically, during the transition of CRF from 0 to 30, our method only experienced a decrease in F1 from 0.545 to 0.504. In particular, our results are consistently higher than 0.5 in all compression cases, which indicates that our SCFNet demonstrates excellent robustness against video compression.

\begin{figure}[!t]
\centering
\includegraphics[width=\linewidth]{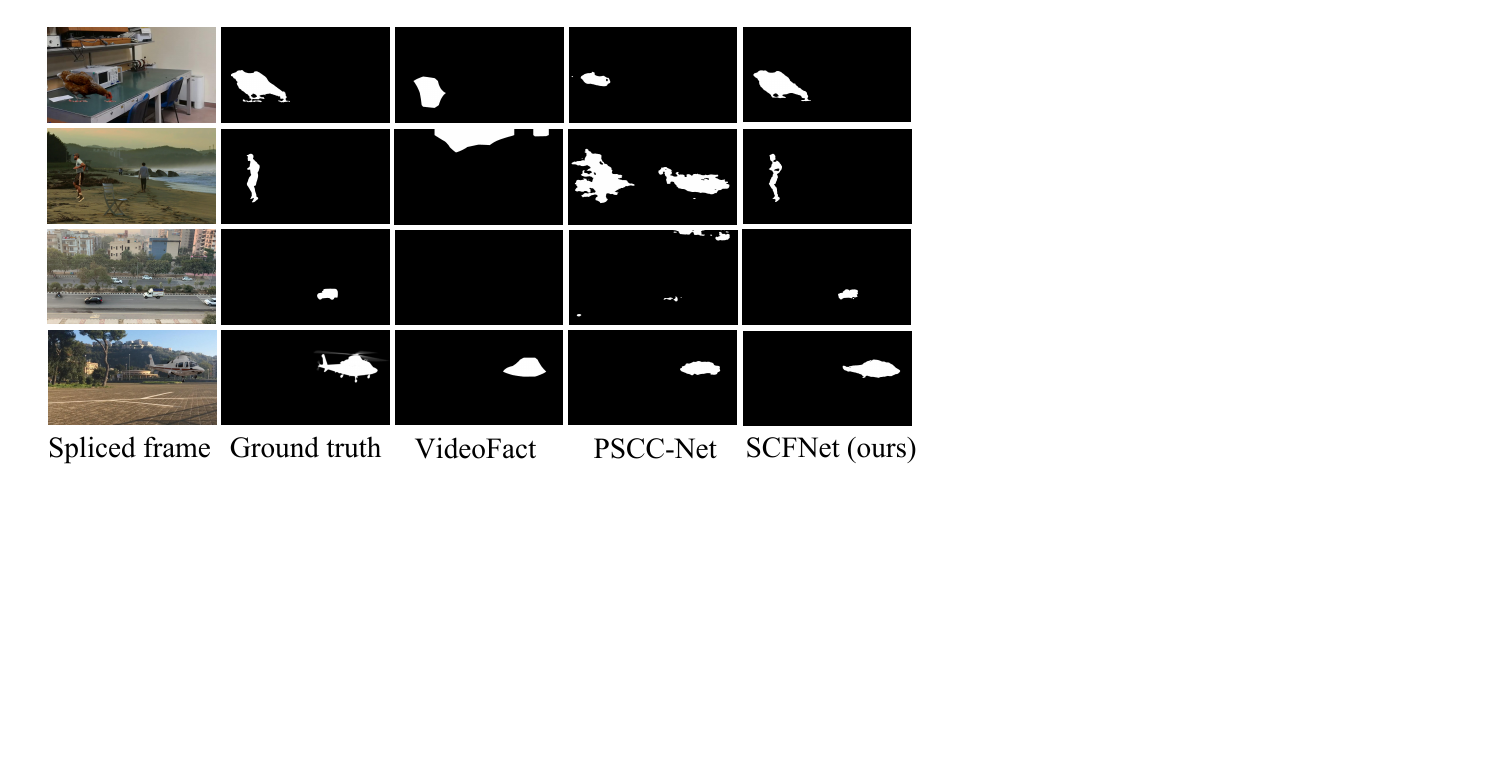}
\vspace{-0.88cm}
\caption{Examples of localization results.}
\label{fig_3}
\end{figure}

\begin{figure}[!t]
\vspace{-0.3cm}
\centering
\includegraphics[width=\linewidth]{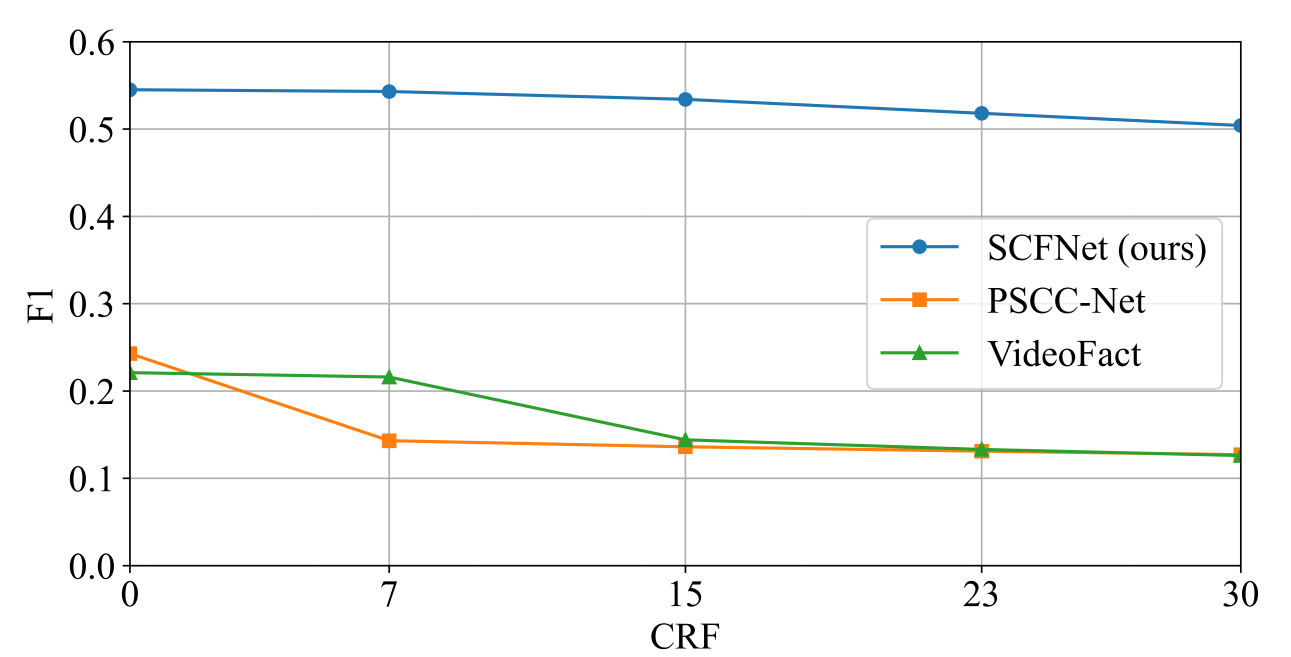}
\vspace{-0.88cm}
\caption{Robustness evaluation results against different compression quality factors (CRFs).}

\label{fig_2}
\end{figure}

\section{CONCLUSION}
In this paper, we propose a novel scheme for video splicing forgery localization. Our method  employs a co-attention fusion based three-stream network to capture manipulation traces present across multiple frames. With the help of co-attention modules and our constructed VSD, the temporal information of splicing videos is thoroughly explored. Extensive experiments demonstrate the superiority of our approach compared with several state-of-the-art methods. In the future work, we aim to expand the proposed method to address intricate video forgery scenarios, including object removal and deepfake videos, etc.

\small

\bibliographystyle{IEEEbib}
\bibliography{strings,refs}

\begin{thebibliography}{10}

\bibitem{richao2014detection}
Richao Chen, Gaobo Yang, and Ningbo Zhu,
\newblock ``Detection of object-based manipulation by the statistical features of object contour,''
\newblock {\em Forensic science international}, vol. 236, pp. 164--169, 2014.

\bibitem{jin2021object}
Xiao Jin, Zhen He, Jing Xu, Yongwei Wang, and Yuting Su,
\newblock ``Object-based video forgery detection via dual-stream networks,''
\newblock in {\em International Conference on Multimedia and Expo}, 2021, pp. 1--6.

\bibitem{nguyen2022videofact}
Tai~D Nguyen, Shengbang Fang, and Matthew~C Stamm,
\newblock ``Videofact: Detecting video forgeries using attention, scene context, and forensic traces,''
\newblock {\em arXiv preprint arXiv:2211.15775}, 2022.

\bibitem{cozzolino2019noiseprint}
Davide Cozzolino and Luisa Verdoliva,
\newblock ``Noiseprint: A cnn-based camera model fingerprint,''
\newblock {\em IEEE Transactions on Information Forensics and Security}, vol. 15, pp. 144--159, 2019.

\bibitem{dong2022mvss}
Chengbo Dong, Xinru Chen, Ruohan Hu, Juan Cao, and Xirong Li,
\newblock ``Mvss-net: Multi-view multi-scale supervised networks for image manipulation detection,''
\newblock {\em IEEE Transactions on Pattern Analysis and Machine Intelligence}, vol. 45, no. 3, pp. 3539--3553, 2022.

\bibitem{liu2022pscc}
Xiaohong Liu, Yaojie Liu, Jun Chen, and Xiaoming Liu,
\newblock ``Pscc-net: Progressive spatio-channel correlation network for image manipulation detection and localization,''
\newblock {\em IEEE Transactions on Circuits and Systems for Video Technology}, vol. 32, no. 11, pp. 7505--7517, 2022.

\bibitem{labartino2013localization}
D~Labartino, Tiziano Bianchi, Alessia De~Rosa, Marco Fontani, David Vazquez-Padin, Alessandro Piva, and Mauro Barni,
\newblock ``Localization of forgeries in mpeg-2 video through gop size and dq analysis,''
\newblock in {\em International Workshop on Multimedia Signal Processing}, 2013, pp. 494--499.

\bibitem{kobayashi2010detecting}
Michihiro Kobayashi, Takahiro Okabe, and Yoichi Sato,
\newblock ``Detecting forgery from static-scene video based on inconsistency in noise level functions,''
\newblock {\em IEEE Transactions on Information Forensics and Security}, vol. 5, no. 4, pp. 883--892, 2010.

\bibitem{kancherla2012novel}
Kesav Kancherla and Srinivas Mukkamala,
\newblock ``Novel blind video forgery detection using markov models on motion residue,''
\newblock in {\em Asian Conference on Intelligent Information and Database Systems}, 2012, pp. 308--315.

\bibitem{d2017autoencoder}
Dario D'avino, Davide Cozzolino, Giovanni Poggi, and Luisa Verdoliva,
\newblock ``Autoencoder with recurrent neural networks for video forgery detection,''
\newblock {\em arXiv preprint arXiv:1708.08754}, 2017.

\bibitem{johnston2020video}
Pamela Johnston, Eyad Elyan, and Chrisina Jayne,
\newblock ``Video tampering localisation using features learned from authentic content,''
\newblock {\em Neural Computing and Applications}, vol. 32, pp. 12243--12257, 2020.

\bibitem{jin2022video}
Xiao Jin, Zhen He, Jing Xu, Yongwei Wang, and Yuting Su,
\newblock ``Video splicing detection and localization based on multi-level deep feature fusion and reinforcement learning,''
\newblock {\em Multimedia Tools and Applications}, vol. 81, no. 28, pp. 40993--41011, 2022.

\bibitem{xie2021segformer}
Enze Xie, Wenhai Wang, Zhiding Yu, Anima Anandkumar, Jose~M Alvarez, and Ping Luo,
\newblock ``Segformer: Simple and efficient design for semantic segmentation with transformers,''
\newblock {\em Advances in Neural Information Processing Systems}, vol. 34, pp. 12077--12090, 2021.

\bibitem{he2016deep}
Kaiming He, Xiangyu Zhang, Shaoqing Ren, and Jian Sun,
\newblock ``Deep residual learning for image recognition,''
\newblock in {\em Computer Vision and Pattern Recognition}, 2016, pp. 770--778.

\bibitem{pei2023hierarchical}
Gensheng Pei, Yazhou Yao, Fumin Shen, Dan Huang, Xingguo Huang, and Heng-Tao Shen,
\newblock ``Hierarchical co-attention propagation network for zero-shot video object segmentation,''
\newblock {\em IEEE Transactions on Image Processing}, 2023.

\bibitem{jaderberg2015spatial}
Max Jaderberg, Karen Simonyan, Andrew Zisserman, et~al.,
\newblock ``Spatial transformer networks,''
\newblock {\em Advances in Neural Information Processing Systems}, vol. 28, 2015.

\bibitem{lu2019see}
Xiankai Lu, Wenguan Wang, Chao Ma, Jianbing Shen, Ling Shao, and Fatih Porikli,
\newblock ``See more, know more: Unsupervised video object segmentation with co-attention siamese networks,''
\newblock in {\em Computer Vision and Pattern Recognition}, 2019, pp. 3623--3632.

\bibitem{xu2018youtube}
Ning Xu, Linjie Yang, Yuchen Fan, Dingcheng Yue, Yuchen Liang, Jianchao Yang, and Thomas Huang,
\newblock ``Youtube-vos: A large-scale video object segmentation benchmark,''
\newblock {\em arXiv preprint arXiv:1809.03327}, 2018.

\bibitem{huang2019got}
Lianghua Huang, Xin Zhao, and Kaiqi Huang,
\newblock ``Got-10k: A large high-diversity benchmark for generic object tracking in the wild,''
\newblock {\em IEEE Transactions on Pattern Analysis and Machine Intelligence}, vol. 43, no. 5, pp. 1562--1577, 2019.

\bibitem{hao2022eiseg}
Yuying Hao, Yi~Liu, Yizhou Chen, Lin Han, Juncai Peng, Shiyu Tang, Guowei Chen, Zewu Wu, Zeyu Chen, and Baohua Lai,
\newblock ``Eiseg: An efficient interactive segmentation annotation tool based on paddlepaddle,''
\newblock {\em arXiv preprint arXiv:2210.08788}, 2022.

\bibitem{cheng2021rethinking}
Ho~Kei Cheng, Yu-Wing Tai, and Chi-Keung Tang,
\newblock ``Rethinking space-time networks with improved memory coverage for efficient video object segmentation,''
\newblock {\em Advances in Neural Information Processing Systems}, vol. 34, pp. 11781--11794, 2021.

\bibitem{cheng2022xmem}
Ho~Kei Cheng and Alexander~G Schwing,
\newblock ``Xmem: Long-term video object segmentation with an atkinson-shiffrin memory model,''
\newblock in {\em European Conference on Computer Vision}, 2022, pp. 640--658.

\bibitem{hosler2019video}
Brian~C Hosler, Xinwei Zhao, Owen Mayer, Chen Chen, James~A Shackleford, and Matthew~C Stamm,
\newblock ``The video authentication and camera identification database: A new database for video forensics,''
\newblock {\em IEEE Access}, vol. 7, pp. 76937--76948, 2019.

\bibitem{singla2023hevc}
Neetu Singla, Jyotsna Singh, Sushama Nagpal, and Bhanu Tokas,
\newblock ``Hevc based tampered video database development for forensic investigation,''
\newblock {\em Multimedia Tools and Applications}, vol. 82, no. 17, pp. 25493--25526, 2023.

\bibitem{mittal2023video}
Trisha Mittal, Ritwik Sinha, Viswanathan Swaminathan, John Collomosse, and Dinesh Manocha,
\newblock ``Video manipulations beyond faces: A dataset with human-machine analysis,''
\newblock in {\em Winter Conference on Applications of Computer Vision}, 2023, pp. 643--652.

\end{thebibliography}

\end{document}